\documentclass[11pt,a4paper]{article}
\usepackage[hyperref]{acl2021}
\usepackage{times}
\usepackage{latexsym}

\usepackage{microtype}

\usepackage{bm}
\usepackage{amsmath}
\usepackage{graphicx}
\usepackage{booktabs}
\usepackage{caption,subcaption}
\usepackage{multirow}
\usepackage{makecell}
\usepackage{enumitem}
\usepackage{amsfonts}
\usepackage{soul}

\usepackage{microtype}
\graphicspath{{./figures/}}
\aclfinalcopy 
\def\aclpaperid{2875} 

\usepackage{color, colortbl}
\definecolor{Gray}{gray}{0.9}

\newcommand{\expnumber}[2]{{#1}\mathrm{e}{#2}}

\title{Meta-Learning with Variational Semantic Memory for Word Sense Disambiguation}

\author{Yingjun Du \\
  University of Amsterdam \\
  \texttt{y.du@uva.nl} \\ \And
  Nithin Holla \\
  Amberscript \\
  \texttt{nithin.holla7@gmail.com} \\ \And
  Xiantong Zhen \\
  University of Amsterdam \\
  \texttt{x.zhen@uva.nl} \\ \AND
  Cees G.M. Snoek \\
  University of Amsterdam \\
  \texttt{C.G.M.Snoek@uva.nl}  \\ \And
  Ekaterina Shutova \\
  University of Amsterdam \\
  \texttt{e.shutova@uva.nl}
  }

\date{}

\begin{document}
\maketitle
\begin{abstract} 
A critical challenge faced by supervised word sense disambiguation (WSD) is the lack of large annotated datasets with sufficient coverage of words in their diversity of senses. This inspired recent research on few-shot WSD using meta-learning. While such work has successfully applied meta-learning to learn new word senses from very few examples, its performance still lags behind its fully-supervised counterpart. Aiming to further close this gap, we propose a model of semantic memory for WSD in a meta-learning setting. Semantic memory encapsulates prior experiences seen throughout the lifetime of the model, which aids better generalization in limited data settings. Our model is based on hierarchical variational inference and incorporates an adaptive memory update rule via a hypernetwork. We show our model advances the state of the art in few-shot WSD, supports effective learning in extremely data scarce (e.g. one-shot) scenarios and produces meaning prototypes that capture similar senses of distinct words.
\end{abstract}

\section{Introduction}

Disambiguating word meaning in context is at the heart of any natural language understanding task or application, whether it is performed explicitly or implicitly. Traditionally, word sense disambiguation (WSD) has been defined as the task of explicitly labeling word usages in context with sense labels from a pre-defined sense inventory. The majority of approaches to WSD rely on (semi\nobreakdash-)supervised learning \citep{yuan-semi_wsd,raganato-framework,raganato-wsd,hadiwinoto-improved_wsd,huang-glossbert,scarlini-wsd, bevilacqua-breaking} and make use of training corpora manually annotated for word senses. Typically, these methods require a fairly large number of annotated training examples per word. This problem is exacerbated by the dramatic imbalances in sense frequencies, which further increase the need for annotation to capture a diversity of senses and to obtain sufficient training data for rare senses.    

This motivated recent research on few-shot WSD, where the objective of the model is to learn new, previously unseen word senses from only a small number of examples. \citet{holla-wsd} presented a meta-learning approach to few-shot WSD, as well as a benchmark for this task. 
Meta-learning makes use of an episodic training regime, where a model is trained on a collection of diverse few-shot tasks and is explicitly optimized to perform well when learning from a small number of examples per task \citep{snell-protonet,finn-maml,triantafillou_metadataset}. \citet{holla-wsd} have shown that meta-learning can be successfully applied to learn new word senses from as little as one example per sense. Yet, the overall model performance in settings where data is highly limited (e.g. one- or two-shot learning) still lags behind that of fully supervised models. 

In the meantime, machine learning research demonstrated the advantages of a memory component for meta-learning in limited data settings  \cite{santoro2016meta,munkhdalai_metanet,munkhdalai2018rapid,VarSemMemory}. The memory stores general knowledge acquired in learning related tasks, which facilitates the acquisition of new concepts and recognition of previously unseen classes with limited labeled data \citep{VarSemMemory}. Inspired by these advances, we introduce the first model of semantic memory for WSD in a meta-learning setting. 
In meta-learning, prototypes are embeddings around which other data points of the same class are clustered \citep{snell-protonet}. Our semantic memory stores prototypical representations of word senses seen during training, 
 generalizing over the contexts in which they are used. This rich contextual information aids in learning new senses of previously unseen words that appear in similar contexts, from very few examples. 

The design of our prototypical representation of word sense takes inspiration from prototype theory \citep{rosch:prototype}, an established account of category representation in psychology. It stipulates that semantic categories are formed around prototypical members, new members are added 
based on resemblance to the prototypes and category membership is a matter of degree. In line with this account, our models learn prototypical representations of word senses from their linguistic context. 
To do this, we employ a neural architecture for learning probabilistic class prototypes: variational prototype networks, augmented with a variational semantic memory (VSM) component \citep{VarSemMemory}. 

Unlike deterministic prototypes in prototypical networks \citep{snell-protonet}, we model class prototypes as distributions and perform variational inference of these prototypes in a hierarchical Bayesian framework. Unlike deterministic memory access in memory-based meta-learning \citep{santoro_mann,munkhdalai_metanet}, we access memory by Monte Carlo sampling from a variational distribution. Specifically, we first perform variational inference to obtain a latent memory variable and then perform another step of variational inference to obtain the prototype distribution. Furthermore, we enhance the memory update of vanilla VSM with a novel adaptive update rule involving a hypernetwork \citep{ha2016hypernetworks} that controls the weight of the updates. We call our approach $\beta$-VSM to denote the adaptive weight $\beta$ for memory updates.


We experimentally demonstrate the effectiveness of this approach for few-shot WSD, advancing the state of the art in this task. Furthermore, we observe the highest performance gains on word senses with the least training examples, emphasizing the benefits of semantic memory for truly few-shot learning scenarios. Our analysis of the meaning prototypes acquired in the memory suggests that they are able to capture related senses of distinct words, demonstrating the generalization capabilities of our memory component. We make our code publicly available to facilitate further research.\footnote{\url{https://github.com/YDU-uva/VSM_WSD}}

\section{Related work}

\paragraph{Word sense disambiguation}

Knowledge-based approaches to WSD \citep{lesk_wsd,agirre-random_walk,moro-babelfy} rely on lexical resources such as WordNet \citep{miller-wordnet} and do not require a corpus manually annotated with word senses. Alternatively, supervised learning methods treat WSD as a word-level classification task for ambiguous words and rely on sense-annotated corpora for training. Early supervised learning approaches trained classifiers with hand-crafted features \citep{navigli_wsd_survey,zhong-IMS} and word embeddings \citep{rothe-autoextend,iacobacci-embeddings_wsd} as input. \citet{raganato-framework} proposed a benchmark for WSD based on the SemCor corpus \citep{miller-semcor} and found that supervised methods outperform the knowledge-based ones. 

Neural models for supervised WSD include LSTM-based \citep{hochreiter_lstm} classifiers \citep{kageback-wsd,melamud-context2vec,raganato-wsd}, nearest neighbour classifier with ELMo embeddings \citep{peters-elmo}, as well as a classifier based on pretrained BERT representations \citep{hadiwinoto-improved_wsd}. Recently, hybrid approaches incorporating information from lexical resources into neural architectures have gained traction. GlossBERT \citep{huang-glossbert} fine-tunes BERT with WordNet sense definitions as additional input. EWISE \citep{kumar-zero_shot_wsd} learns continuous sense embeddings as targets, aided by dictionary definitions and lexical knowledge bases. \citet{scarlini-wsd} present a semi-supervised approach for obtaining sense embeddings with the aid of a lexical knowledge base, enabling WSD with a nearest neighbor algorithm. By further exploiting the graph structure of WordNet and integrating it with BERT, EWISER \citep{bevilacqua-breaking} achieves the current state-of-the-art performance on the benchmark by \citet{raganato-framework} -- an F1 score of $80.1\%$.

Unlike few-shot WSD, these works do not fine-tune the models on new words during testing. Instead, they train on a training set and evaluate on a test set where words and senses might have been seen during training.

\paragraph{Meta-learning}
Meta-learning, or learning to learn \citep{schmidhuber_thesis,bengio_metalearning,thrun_metalearning}, is a learning paradigm where a model is trained on a distribution of tasks so as to enable rapid learning on new tasks. By solving a large number of different tasks, it aims to leverage the acquired knowledge to learn new, unseen tasks. The training set, referred to as the \textit{meta-training set}, consists of \textit{episodes}, each corresponding to a distinct task. Every episode is further divided into a \textit{support set} containing just a handful of examples for learning the task, and a \textit{query set} containing examples for task evaluation. In the meta-training phase, for each episode, the model adapts to the task using the support set, and its performance on the task is evaluated on the corresponding query set. The initial parameters of the model are then adjusted based on the loss on the query set. By repeating the process on several episodes/tasks, the model produces representations that enable rapid adaptation to a new task. The test set, referred to as the \textit{meta-test set}, also consists of episodes with a support and query set. The meta-test set corresponds to new tasks that were not seen during meta-training. During meta-testing, the meta-trained model is first fine-tuned on a small number of examples in the support set of each meta-test episode and then evaluated on the accompanying query set. The average performance on all such query sets measures the few-shot learning ability of the model.

Metric-based meta-learning methods \citep{Koch2015SiameseNN,Vinyals,sung2017_relationalnet,snell-protonet} learn a kernel function and make predictions on the query set based on the similarity with the support set examples. Model-based methods \citep{santoro_mann,munkhdalai_metanet} employ external memory and make predictions based on examples retrieved from the memory. Optimization-based methods \citep{Ravi,finn-maml,Nichol,antoniou-maml_plus} directly optimize for generalizability over tasks in their training objective. 

Meta-learning has been applied to a range of tasks in NLP, including machine translation \citep{gu-etal-2018-meta}, relation classification \citep{obamuyide-vlachos-2019-rel-classification}, text classification \citep{yu-diverse,geng-induction}, hypernymy detection \citep{yu-hypernymy}, and dialog generation \citep{qian-dialog}. It has also been used to learn across distinct NLP tasks \citep{dou-etal-meta,bansal_meta} as well as across different languages \citep{nooralahzadeh-crosslingual,li-metagraph}.
\citet{bansal-self} show that meta-learning during self-supervised pretraining of language models leads to improved few-shot generalization on downstream tasks. 

\citet{holla-wsd} propose a framework for few-shot word sense disambiguation, where the goal is to disambiguate new words during meta-testing. Meta-training consists of episodes formed from multiple words whereas meta-testing has one episode corresponding to each of the test words. They show that prototype-based methods -- prototypical networks \citep{snell-protonet} and first-order ProtoMAML \citep{triantafillou_metadataset} -- obtain promising results, in contrast with model-agnostic meta-learning (MAML) \citep{finn-maml}.

\paragraph{Memory-based models}
Memory mechanisms \cite{weston2014memory,graves_nmt,krotov2016dense} have recently drawn increasing attention. In memory-augmented neural network \citep{santoro_mann}, given an input, the memory read and write operations are performed by a controller, using soft attention for reads and least recently used access module for writes. Meta Network \citep{munkhdalai-metanet} uses two memory modules: a key-value memory in combination with slow and fast weights for one-shot learning. An external memory was introduced to enhance recurrent neural network in \citet{munkhdalai2019metalearned}, in which memory is conceptualized as an adaptable function and implemented as a deep neural network. Semantic memory has recently been introduced by \citet{VarSemMemory} for few-shot learning to enhance prototypical representations of objects, where memory recall is cast as a variational inference problem.


In NLP, \citet{tang-aspect} use content and location-based neural attention over external memory for aspect-level sentiment classification. \citet{das-qa} use key-value memory for question answering on knowledge bases. Mem2Seq \citep{madotto-mem2seq} is an architecture for task-oriented dialog that combines attention-based memory with pointer networks \citep{vinyals-pointer}.  \citet{geng-dynamic} propose Dynamic Memory Induction Networks for few-shot text classification, which utilizes dynamic routing \citep{sabour_capsule} over a static memory module. Episodic memory has been used in lifelong learning on language tasks, as a means to perform experience replay \citep{deMasson-episodic_memory,han-continual,holla-lifelong}.  

\section{Task and dataset}


We treat WSD as a word-level classification problem where ambiguous words are to be classified into their senses given the context. In traditional WSD, the goal is to generalize to new contexts of word-sense pairs. Specifically, the test set consists of word-sense pairs that were seen during training. On the other hand, in few-shot WSD, the goal is to generalize to new words and senses altogether. The meta-testing phase involves further adapting the models (on the small support set) to new words that were not seen during training and evaluates them on new contexts (using the query set). It deviates from the standard $N$-way, $K$-shot classification setting in few-shot learning since the words may have a different number of senses and each sense may have different number of examples \citep{holla-wsd}, making it a more realistic few-shot learning setup \citep{triantafillou_metadataset}. 

\paragraph{Dataset} We use the few-shot WSD benchmark provided by \citet{holla-wsd}. It is based on the SemCor corpus \citep{miller-semcor}, annotated with senses from the New Oxford American Dictionary by \citet{yuan-semi_wsd}. The dataset consists of words grouped into meta-training, meta-validation and meta-test sets. The meta-test set consists of new words that were not part of meta-training and meta-validation sets. There are four setups varying in the number of sentences in the support set $|S| = 4, 8, 16, 32$. $|S| = 4$ corresponds to an extreme few-shot learning scenario for most words, whereas $|S| = 32$ comes closer to the number of sentences per word encountered in standard WSD setups. For $|S| = 4, 8, 16, 32$, the number of unique words in the meta-training / meta-validation / meta-test sets is 985/166/270, 985/163/259, 799/146/197 and 580/85/129 respectively. We use the publicly available standard dataset splits.\footnote{\url{https://github.com/Nithin-Holla/MetaWSD}}

\paragraph{Episodes} The meta-training episodes were created by first sampling a set of words and a fixed number of senses per word, followed by sampling example sentences for these word-sense pairs. This strategy allows for a combinatorially large number of episodes. Every meta-training episode has $|S|$ sentences in both the support and query sets, and corresponds to the distinct task of disambiguating between the sampled word-sense pairs. The total number of meta-training episodes is $10,000$. In the meta-validation and meta-test sets, each episode corresponds to the task of disambiguating a single, previously unseen word between all its senses. For every meta-test episode, the model is fine-tuned on a few examples in the support set and its generalizability is evaluated on the query set. In contrast to the meta-training episodes, the meta-test episodes reflect a natural distribution of senses in the corpus, including class imbalance, providing a realistic evaluation setting.

\section{Methods}

\subsection{Model architectures}
We experiment with the same model architectures as \citet{holla-wsd}. The model $f_{\theta}$, with parameters $\theta$, takes words $\mathbf{x}_i$ as input and produces a per-word representation vector $f_{\theta}(\mathbf{x}_i)$ for $ i = 1, ..., L$ where $L$ is the length of the sentence. Sense predictions are only made for ambiguous words using the corresponding word representation. 

\paragraph{GloVe+GRU} Single-layer bi-directional GRU \citep{cho-gru} network followed by a single linear layer, that takes GloVe embeddings \citep{pennington-glove} as input. GloVe embeddings capture all senses of a word. We thus evaluate a model's ability to disambiguate from sense-agnostic input. 

\paragraph{ELMo+MLP} A multi-layer perception (MLP) network that receives contextualized ELMo embeddings \citep{peters-elmo} as input. Their contextualised nature makes ELMo embeddings better suited to capture meaning variation than the static ones. Since ELMo is not fine-tuned, this model has the lowest number of learnable parameters.

\paragraph{BERT} Pretrained BERT\textsubscript{BASE} \citep{devlin-bert} model followed by a linear layer, fully fine-tuned on the task. BERT underlies state-of-the-art approaches to WSD.

\subsection{Prototypical Network}

Our few-shot learning approach builds upon prototypical networks~\citep{snell-protonet}, which is widely used for few-shot image classification and has been shown to be successful in WSD \citep{holla-wsd}. It computes a prototype $\mathbf{z}_k = \frac{1}{K}\sum_k f_{\theta}(\mathbf{x}_k)$ of each word sense (where $K$ is the number of examples for each word sense) through an embedding function $f_{\theta}$, which is realized as the aforementioned architectures. It computes a distribution over classes for a query sample $\mathbf{x}$ given a distance function $d(\cdot, \cdot)$ as the softmax over its distances to the prototypes in the embedding space:
\begin{equation}
    p(\mathbf{y}_{i} = k|\mathbf{x}) = \frac{\exp(-d(f_{\theta}(\mathbf{x}),\mathbf{z}_k))}{\sum_{k'} \exp(-d(f_{\theta}(\mathbf{x}),\mathbf{z}_{k'}))}
    \label{eq:prototype}
\end{equation}

However, the resulting prototypes may not be sufficiently representative of word senses as semantic categories when using a single deterministic vector, computed as the average of only a few examples. Such representations lack expressiveness and may not encompass sufficient intra-class variance, that is needed to distinguish between different fine-grained word senses. Moreover, large uncertainty arises in the single prototype due to the small number of samples. 

\subsection{Variational Prototype Network}

Variational prototype network~\citep{VarSemMemory} (VPN) is a powerful model for learning latent representations from small amounts of data, where the prototype $\mathbf{z}$ of each class is treated as a distribution. Given a task with a support set $S$ and query set $Q$, the objective of VPN takes the following form:
\begin{equation}
\begin{aligned}
   \mathcal{L}_{\mathrm{VPN}} &= \frac{1}{|Q|} \sum^{|Q|}_{i=1} \Big[\frac{1}{L_\mathbf{z}} \sum_{l_\mathbf{z}=1}^{L_\mathbf{z}} -\log p(\mathbf{y}_i|\mathbf{x}_i,\mathbf{z}^{(l_\mathbf{z})})
    \\& + \lambda D_{\mathrm{KL}}[q(\mathbf{z}|S)||p(\mathbf{z}|\mathbf{x}_i)]\Big]
\label{L_VPN}    
\end{aligned}
\end{equation}
where $q(\mathbf{z}|S)$ is the variational posterior over $\mathbf{z}$, $p(\mathbf{z}|\mathbf{x}_i)$ is the prior, and $L_\mathbf{z}$ is the number of Monte Carlo samples for $\mathbf{z}$. The prior and posterior are assumed to be Gaussian.  The re-parameterization trick~\citep{kingma2013auto} is adopted to enable back-propagation with gradient descent, i.e., $\mathbf{z}^{(l_\mathbf{z})} = f(S, \epsilon^{(l_\mathbf{z})})$, $\epsilon^{(l_\mathbf{z})} \sim \mathcal{N} (0, I)$, $f(\cdot, \cdot) = \epsilon^{(l_\mathbf{z})} * \mu_z + \sigma_z$, where the mean $\mu_z$ and diagonal covariance $\sigma_z$ are generated from the posterior inference network with $S$ as input.
The amortization technique is employed for the implementation of VPN. The posterior network takes the mean word representations in the support set $S$ as input and returns the parameters of $q(\mathbf{z}|S)$. Similarly, the prior network produces the parameters of $p(\mathbf{z}|\mathbf{x}_i)$ by taking the query word representation $\mathbf{x_i} \in \mathcal{Q}$ as input. The conditional predictive log-likelihood is implemented as a cross-entropy loss. 


\subsection{$\beta$-Variational Semantic Memory}

\begin{figure}[t]
\centering
  \includegraphics[width=.8\linewidth]{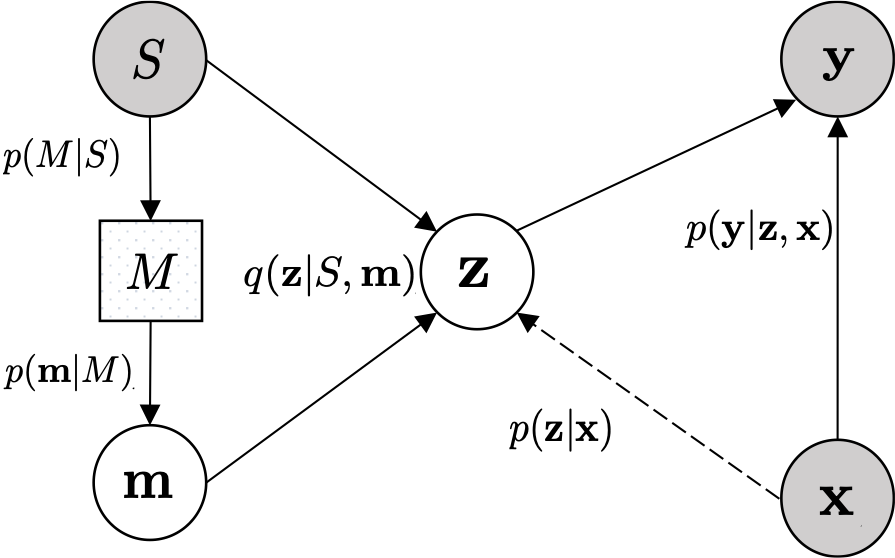}
 \caption{Computational graph of  variational semantic memory for few-shot WSD. $M$ is the semantic memory module, $S$ the support set, $\mathbf{x}$ and $\mathbf{y}$ are the query sample and label, and $\mathbf{z}$ is the word sense prototype.}
 \label{fig:framework}
\end{figure}

In order to leverage the shared common knowledge between different tasks to improve disambiguation in future tasks, we incorporate variational semantic memory (VSM) as in \citet{VarSemMemory}. 
It consists of two main processes: \textit{memory recall}, which retrieves relevant information that fits with specific tasks based on the support set of the current task; \textit{memory update}, which  effectively collects new information from the task and gradually consolidates the semantic knowledge in the memory. We adopt a similar memory mechanism and introduce an improved update rule for memory consolidation.

\paragraph{Memory recall} The memory recall of VSM aims to choose the related content from the memory, and is accomplished by variational inference. It introduces latent memory $\mathbf{m}$ as an intermediate stochastic variable, and infers $\mathbf{m}$ from the addressed memory $M$. The approximate variational posterior $q(\mathbf{m}|M, S)$ over the latent memory $\mathbf{m}$ is obtained empirically by

\begin{equation}
q(\mathbf{m}|M,S) = \sum^{|M|}_{a=1}\gamma_a p(\mathbf{m}|M_a), 
\label{qms}
\end{equation}
where 
\begin{equation}
\gamma_a = \frac{\exp\big(g(M_a,S)\big)}{\sum_i \exp\big(g(M_i,S)\big)}
\label{lambda}
\end{equation}
$g(\cdot)$ is the dot product, $|M|$ is the number of memory slots, $M_a$ is the memory content at slot $a$ and stores the prototype of samples in each class, and we take the mean representation of samples in $S$. 

The variational posterior over the prototype then becomes:
\begin{equation}
    \tilde{q}(\mathbf{z}|M,S) \approx \frac{1}{L_{\mathbf{m}}}\sum^{L_m}_{l_{\mathbf{m}}=1} q(\mathbf{z}|\mathbf{m}^{(l_{\mathbf{m}})},S),
    \label{}
\end{equation}
where $\mathbf{m}^{(l_{\mathbf{m}})}$ is a Monte Carlo sample drawn from the distribution $q(\mathbf{m}|M,S)$, and $l_{\mathbf{m}}$ is the number of samples. By incorporating the latent memory $\mathbf{m}$ from Eq.~(\ref{qms}), we achieve the objective for variational semantic memory as follows:
\begin{equation}
\begin{aligned}
    \mathcal{L}_{\rm{VSM}} &= \sum^{|Q|}_{i=1} \Big[ -\mathbb{E}_{q(\mathbf{z}|S, \mathbf{m})} \big[\log p(\mathbf{y}_i|\mathbf{x}_i,\mathbf{z})\big]  
    \\& + \lambda_{\mathbf{z}} D_{\mathrm{KL}}\big[q(\mathbf{z}|S, \mathbf{m})||p(\mathbf{z}|\mathbf{x}_i)\big]
      \\& + \lambda_{\mathbf{m}} D_{\mathrm{KL}}\big[\sum^{|M|}_{i}\gamma_i p(\mathbf{m}|M_i)||p(\mathbf{m}|S)\big]\Big] 
    \label{obj}
\end{aligned}
\end{equation}
where $p(\mathbf{m}|S)$ is the introduced prior over $\mathbf{m}$, $\lambda_{\mathbf{z}}$ and $\lambda_{\mathbf{m}}$ are the hyperparameters.
The overall  computational graph of VSM is shown in Figure~\ref{fig:framework}. Similarly, the posterior and prior over $\mathbf{m}$ are also assumed to be Gaussian and obtained by using amortized inference networks; more details are provided in Appendix \ref{sec:implementation}.

\paragraph{Memory update} The memory update is to be able to effectively absorb new useful information to enrich memory content. 
VSM employs an update rule as follows:
%
\begin{equation}
    M_c \leftarrow \beta M_c + (1-\beta) \bar{M}_c,
\end{equation}
where $M_c$ is the memory content corresponding to class $c$, $\bar{M}_c$ is obtained using graph attention~\citep{velivckovic2017graph}, and $\beta \in (0,1)$ is a hyperparameter.

\paragraph{Adaptive memory update} Although VSM was shown to be promising for few-shot image classification, it can be seen from the experiments by \citet{VarSemMemory} that different values of $\beta$ have considerable influence on the performance. $\beta$ determines the extent to which memory is updated at each iteration. In the original VSM, $\beta$ is treated as a hyperparameter obtained by cross-validation, which is time-consuming and inflexible in dealing with different datasets. 
To address this problem, we propose an adaptive memory update rule by learning $\beta$ from data using a lightweight hypernetwork~\citep{ha2016hypernetworks}. 
To be more specific, we obtain $\beta$ by a function $f_{\beta}(\cdot)$ implemented as an MLP with a sigmoid activation function in the output layer. 
The hypernetwork takes $\bar{M}_c$ as input and returns the value of $\beta$:
\begin{equation} \label{eqn:hyp_net}
   \beta = f_{\beta}(\bar{M}_c)
\end{equation}
Moreover, to prevent the possibility of endless growth of memory value, we propose to scale down the memory value whenever $\left \|M_c \right\|_2 > 1$. This is achieved by scaling as follows:

\begin{equation}
    M_c = \frac{M_c}{\max (1, \left \|M_c \right\|_2 )}
\end{equation}

When we update memory, we feed the new obtained memory $\bar{M}_c$ into the hypernetwork $f_{\beta}(\cdot)$ and output adaptive $\beta$ for the update. We provide a more detailed implementation of $\beta$-VSM in Appendix \ref{sec:implementation}.

\section{Experiments and results}

\begin{table*}[ht]
\small
\centering
\begin{tabular}{llllll}
\toprule
\multirow{2}{*}{\makecell{\textbf{Embedding/} \\ \textbf{Encoder}}} & \multicolumn{1}{c}{\multirow{2}{*}{\textbf{Method}}} &  \multicolumn{4}{c}{\textbf{Average macro F1 score}} \\
                       & & \multicolumn{1}{c}{$|S| = 4$} & \multicolumn{1}{c}{$|S| = 8$}  & \multicolumn{1}{c}{$|S| = 16$} & \multicolumn{1}{c}{$|S| = 32$} \\ \midrule
- & MajoritySenseBaseline & 0.247 & 0.259 & 0.264 & 0.261 \\ \midrule
\multirow{5}{*}{GloVe+GRU} & NearestNeighbor & -- & -- & -- & -- \\
& EF-ProtoNet & 0.522 $\pm$ 0.008 & 0.539 $\pm$ 0.009 & 0.538 $\pm$ 0.003 & 0.562 $\pm$ 0.005 \\
& ProtoNet & 0.579 $\pm$ 0.004 & 0.601 $\pm$ 0.003 & 0.633 $\pm$ 0.008 & 0.654 $\pm$ 0.004 \\
& ProtoFOMAML & 0.577 $\pm$ 0.011 & 0.616 $\pm$ 0.005 & 0.626 $\pm$ 0.005 & 0.631 $\pm$ 0.008 \\
\rowcolor{Gray} 
& $\beta$-VSM (Ours) & \textbf{0.597 $\pm$ 0.005} & \textbf{0.631 $\pm$ 0.004} & \textbf{0.652 $\pm$ 0.006} & \textbf{0.678 $\pm$ 0.007} \\
\midrule
\multirow{5}{*}{ELMo+MLP} & NearestNeighbor & 0.624 & 0.641 & 0.645 & 0.654 \\
& EF-ProtoNet & 0.609 $\pm$ 0.008 & 0.635 $\pm$ 0.004 & 0.661 $\pm$ 0.004 &  0.683 $\pm$ 0.003 \\
& ProtoNet & 0.656 $\pm$ 0.006 & 0.688 $\pm$ 0.004 & 0.709 $\pm$ 0.006 & 0.731 $\pm$ 0.006 \\
& ProtoFOMAML & 0.670 $\pm$ 0.005 & 0.700 $\pm$ 0.004 & 0.724 $\pm$ 0.003 & 0.737 $\pm$ 0.007 \\
\rowcolor{Gray} 
& $\beta$-VSM (Ours)  & \textbf{0.679 $\pm$ 0.006} & \textbf{0.709 $\pm$ 0.005} & \textbf{0.735 $\pm$ 0.004} & \textbf{0.758 $\pm$ 0.005}  \\ \midrule
\multirow{5}{*}{BERT} & NearestNeighbor & 0.681 & 0.704 & 0.716 & 0.741 \\
& EF-ProtoNet & 0.594 $\pm$ 0.008 & 0.655 $\pm$ 0.004 & 0.682 $\pm$ 0.005 & 0.721 $\pm$ 0.009 \\
& ProtoNet & 0.696 $\pm$ 0.011 & 0.750 $\pm$ 0.008 & 0.755 $\pm$ 0.002 & 0.766 $\pm$ 0.003 \\
& ProtoFOMAML & 0.719 $\pm$ 0.005 & 0.756 $\pm$ 0.007 & 0.744 $\pm$ 0.007 & 0.761 $\pm$ 0.005 \\
\rowcolor{Gray} 
& $\beta$-VSM (Ours) & \textbf{0.728 $\pm$ 0.012} & \textbf{0.773 $\pm$ 0.005} & \textbf{0.776 $\pm$ 0.003} & \textbf{0.788 $\pm$ 0.003} \\
\bottomrule                       
\end{tabular}
\caption{Model performance comparison on the meta-test words using different embedding functions. 
}
\label{tab:results}
\end{table*}

\paragraph{Experimental setup}
The size of the shared linear layer and memory content of each word sense is 64, 256, and 192 for GloVe+GRU, ELMo+MLP and BERT respectively. The activation function of the shared linear layer is tanh for GloVe+GRU and ReLU for the rest.  The inference networks $g_{\phi}(\cdot)$ for calculating the prototype distribution  and $g_{\psi}(\cdot)$ for calculating the memory distribution are all three-layer MLPs, with the size of each hidden layer being 64, 256, and 192 for GloVe+GRU, ELMo+MLP and BERT. The activation function of their hidden layers is ELU \citep{clevert-elu}, and the output layer does not use any activation function. Each batch during meta-training includes 16 tasks. The hypernetwork $f_{\beta}(\cdot)$ is also a three-layer MLP, with the size of hidden state consistent with that of the memory contents. The linear layer activation function is ReLU for the hypernetwork. 
For BERT and $|S| = \{4,8\}$,  $\lambda_\mathbf{z} = 0.001$, $\lambda_\mathbf{m} = 0.0001$ and learning rate is $\expnumber{5}{-6}$; $|S| = 16$, $\lambda_\mathbf{z} = 0.0001$, $\lambda_\mathbf{m} = 0.0001$ and learning rate is $\expnumber{1}{-6}$; $|S| = 32$, $\lambda_\mathbf{z} = 0.001$, $\lambda_\mathbf{m} = 0.0001$ and learning rate is $\expnumber{1}{-5}$. Hyperparameters for other models are reported in Appendix \ref{sec:hyperparameters}.
All the hyperparameters are chosen using the meta-validation set. The number of slots in memory is consistent with the number of senses in the meta-training set -- $2915$ for $|S| = 4$ and $8$; $2452$ for $|S| = 16$; $1937$ for $|S| = 32$. The evaluation metric is the word-level macro F1 score, averaged over all episodes in the meta-test set. The parameters are optimized using Adam~\citep{kingma2014adam}.

We compare our methods against several baselines and state-of-the-art approaches. 
The nearest neighbor classifier baseline (\textit{NearestNeighbor}) predicts a query example's sense as the sense of the support example closest in the word embedding space (ELMo and BERT) in terms of cosine distance. The episodic fine-tuning baseline (\textit{EF-ProtoNet}) is one where only meta-testing is performed, starting from a randomly initialized model. Prototypical network (\textit{ProtoNet}) and ProtoFOMAML achieve the highest few-shot WSD performance to date on the benchmark of \citet{holla-wsd}. 

\paragraph{Results}
In Table~\ref{tab:results}, we show the average macro F1 scores of the models, with their mean and standard deviation obtained over five independent runs. Our proposed $\beta$-VSM achieves the new state-of-the-art performance on few-shot WSD with all the embedding functions, across all the setups with varying $|S|$. For GloVe+GRU, where the input is sense-agnostic embeddings, our model improves disambiguation compared to ProtoNet by $1.8\%$ for $|S| = 4$ and by $2.4\%$ for $|S| = 32$. With contextual embeddings as input, $\beta$-VSM with ELMo+MLP also leads to improvements compared to the previous best ProtoFOMAML for all $|S|$. \citet{holla-wsd} obtained state-of-the-art performance with BERT, and $\beta$-VSM further advances this, resulting in a gain of $0.9$ -- $2.2\%$. The consistent improvements with different embedding functions and support set sizes suggest that our $\beta$-VSM is effective for few-shot WSD for varying number of shots and senses as well as across model architectures.

\begin{table*}[ht]
\small
\centering
\begin{tabular}{llllll}
\toprule
\multirow{2}{*}{\makecell{\textbf{Embedding/} \\ \textbf{Encoder}}} & \multicolumn{1}{c}{\multirow{2}{*}{\textbf{Method}}} &  \multicolumn{4}{c}{\textbf{Average macro F1 score}} \\
                       & & \multicolumn{1}{c}{$|S| = 4$} & \multicolumn{1}{c}{$|S| = 8$}  & \multicolumn{1}{c}{$|S| = 16$} & \multicolumn{1}{c}{$|S| = 32$} \\ \midrule
\multirow{4}{*}{GloVe+GRU} 
& ProtoNet & 0.579 $\pm$ 0.004 & 0.601 $\pm$ 0.003 & 0.633 $\pm$ 0.008 & 0.654 $\pm$ 0.004 \\
& VPN& 0.583 $\pm$ 0.005 & 0.618 $\pm$ 0.005 & 0.641 $\pm$ 0.007 & 0.668 $\pm$ 0.005 \\
& VSM& 0.587 $\pm$ 0.004 & 0.625 $\pm$ 0.004 & 0.645 $\pm$ 0.006 & 0.670 $\pm$ 0.005 \\
& $\beta$-VSM  & \textbf{0.597 $\pm$ 0.005} & \textbf{0.631 $\pm$ 0.004} & \textbf{0.652 $\pm$ 0.006} & \textbf{0.678 $\pm$ 0.007} \\
\midrule
\multirow{4}{*}{ELMo+MLP} 
& ProtoNet & 0.656 $\pm$ 0.006 & 0.688 $\pm$ 0.004 & 0.709 $\pm$ 0.006 & 0.731 $\pm$ 0.006 \\
& VPN & 0.661 $\pm$ 0.005 & 0.694 $\pm$ 0.006 & 0.718 $\pm$ 0.004 & 0.741 $\pm$ 0.004  \\ 
& VSM& 0.670 $\pm$ 0.006 & 0.707 $\pm$ 0.006 & 0.726 $\pm$ 0.005 & 0.750 $\pm$ 0.004  \\ 
& $\beta$-VSM & \textbf{0.679 $\pm$ 0.006} & \textbf{0.709 $\pm$ 0.005} & \textbf{0.735 $\pm$ 0.004} & \textbf{0.758 $\pm$ 0.005}  \\
\midrule
\multirow{4}{*}{BERT} 
& ProtoNet & 0.696 $\pm$ 0.011 & 0.750 $\pm$ 0.008 & 0.755 $\pm$ 0.002 & 0.766 $\pm$ 0.003 \\
& VPN & 0.703 $\pm$ 0.011 & 0.761 $\pm$ 0.007 & 0.762 $\pm$ 0.004 & 0.779 $\pm$ 0.002 \\
& VSM& 0.717 $\pm$ 0.013 & 0.769 $\pm$ 0.006 & 0.770 $\pm$ 0.005 & 0.784 $\pm$ 0.002 \\
& $\beta$-VSM  & \textbf{0.728 $\pm$ 0.012} & \textbf{0.773 $\pm$ 0.005} & \textbf{0.776 $\pm$ 0.003} & \textbf{0.788 $\pm$ 0.003} \\
\bottomrule                       
\end{tabular}
\caption{Ablation study comparing the meta-test performance of the different variants of prototypical networks.}
\label{tab:vpn}
\end{table*}

\begin{figure*}[ht]
    \centering
    \begin{subfigure}[b]{0.24\textwidth}
         \centering
         \includegraphics[width=\textwidth]{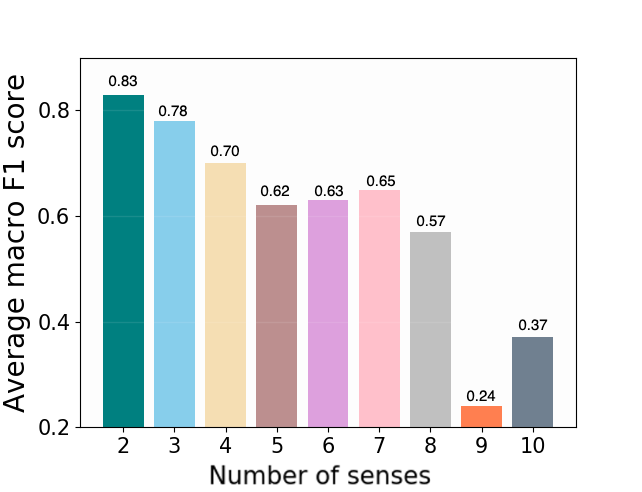}
         \caption{ProtoNet}
         \label{fig:f1_hist_4}
     \end{subfigure}
     \hfill
    \begin{subfigure}[b]{0.24\textwidth}
         \centering
         \includegraphics[width=\textwidth]{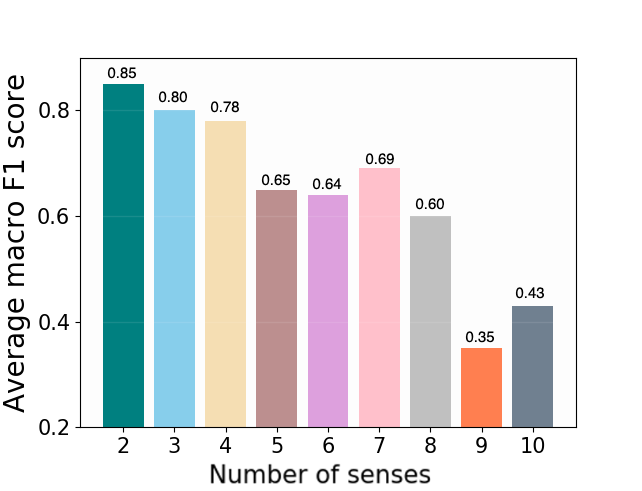}
         \caption{VPN}
         \label{fig:VPN_8}
     \end{subfigure}
     \hfill
     \begin{subfigure}[b]{0.24\textwidth}
         \centering
         \includegraphics[width=\textwidth]{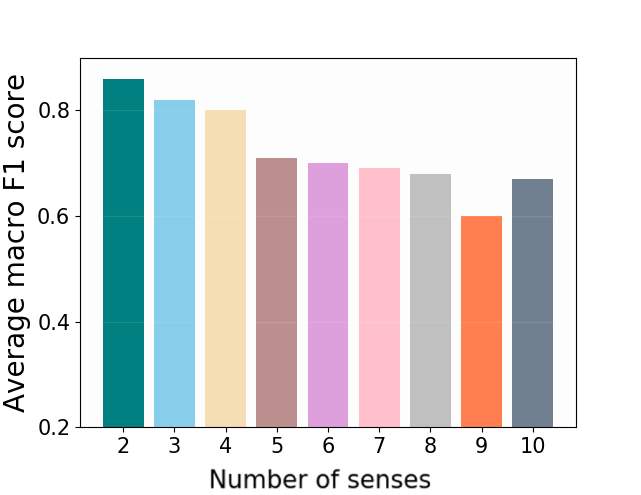}
         \caption{VSM}
         \label{fig:VSM_16}
     \end{subfigure}
     \hfill
     \begin{subfigure}[b]{0.24\textwidth}
         \centering
         \includegraphics[width=\textwidth]{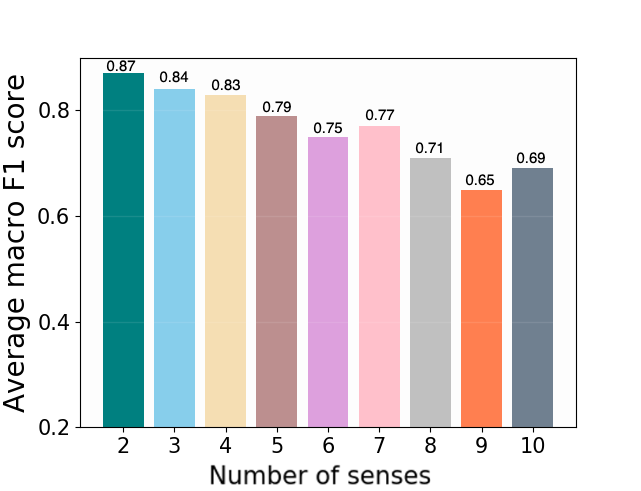}
         \caption{$\beta$-VSM}
         \label{fig:f1_hist_32}
     \end{subfigure}
    \caption{Distribution of average macro F1 scores over number of senses for BERT-based models with $|S| = 16$.}
    \label{fig:f1_hist}
\end{figure*}

\begin{figure*}[t]
    \centering
    \begin{subfigure}[b]{0.3\textwidth}
         \centering
         \includegraphics[width=\textwidth]{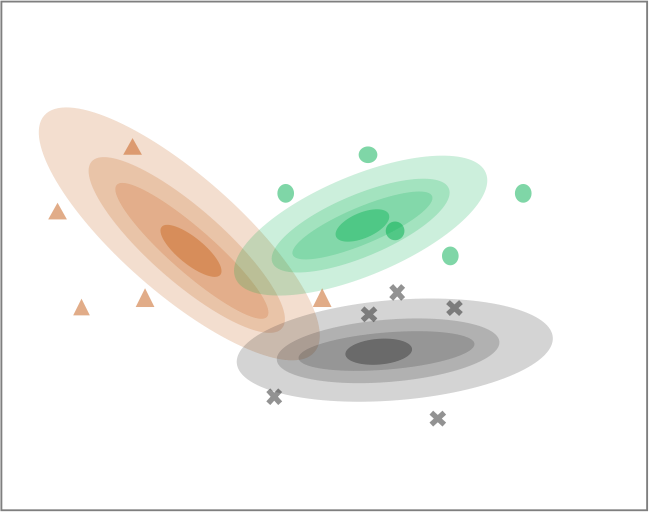}
         \caption{VPN}
         \label{fig:vpn_tsne}
     \end{subfigure}
     \hfill
    \begin{subfigure}[b]{0.3\textwidth}
         \centering
         \includegraphics[width=\textwidth]{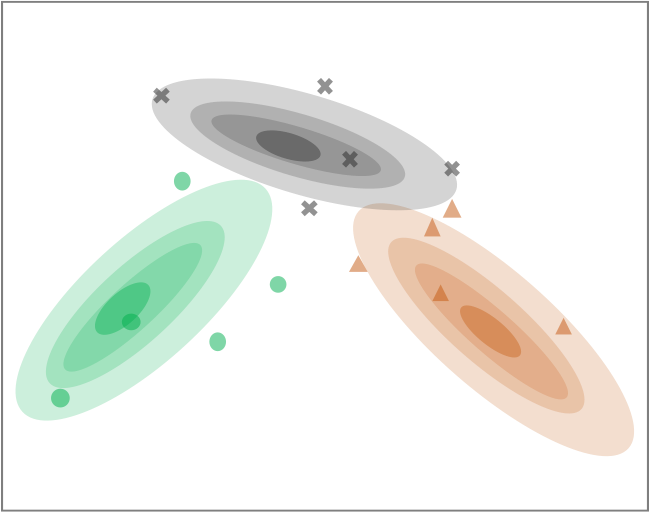}
         \caption{VSM}
         \label{fig:VSM_tsne}
     \end{subfigure}
     \hfill
     \begin{subfigure}[b]{0.3\textwidth}
         \centering
         \includegraphics[width=\textwidth]{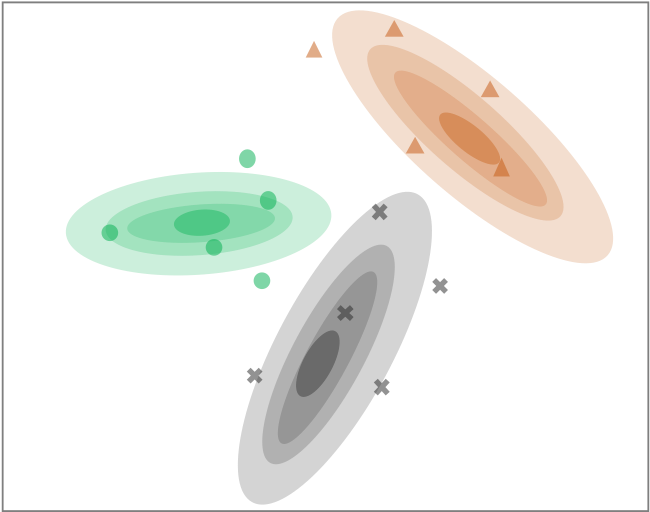}
         \caption{$\beta$-VSM}
         \label{fig:beta_tsne}
     \end{subfigure}
    \caption{Prototype distributions of distinct senses of \textit{draw} with different models.}
    \label{fig:tsne_1}
\end{figure*}

\begin{figure}[ht]
\centering
  \includegraphics[width=.65\linewidth]{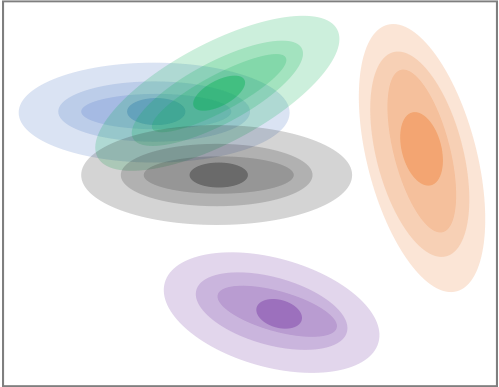}
 \caption{Prototype distributions of similar sense of \textit{launch} (blue), \textit{start} (green) and \textit{establish} (grey). Distinct senses: \textit{start} (orange) and \textit{establish} (purple).  }
 \label{fig:memory}
\end{figure}

\section{Analysis and discussion}

To analyze the contributions of different components in our method, we perform an ablation study by comparing ProtoNet, VPN, VSM and $\beta$-VSM and present the macro F1 scores in Table~\ref{tab:vpn}.

\paragraph{Role of variational prototypes} 
VPN consistently outperforms ProtoNet with all embedding functions (by around 1\% F1 score on average). 
The results indicate that the probabilistic prototypes provide more informative representations of word senses compared to deterministic vectors. The highest gains were obtained in case of GloVe+GRU ($1.7\%$ F1 score with $|S| = 8$), suggesting that probabilistic prototypes are particularly useful for models that rely on static word embeddings, as they capture uncertainty in contextual interpretation.

\paragraph{Role of variational semantic memory}
We show the benefit of VSM by comparing it with VPN. VSM consistently surpasses VPN with all three embedding functions. 
According to our analysis, VSM makes the prototypes of different word senses more distinctive and distant from each other. The senses in memory provide more context information, enabling larger intra-class variations to be captured, and thus lead to improvements upon VPN. 


\paragraph{Role of adaptive $\beta$}
To demonstrate the effectiveness of the hypernetwork for adaptive $\beta$, we compare $\beta$-VSM with VSM where  $\beta$ is tuned by cross-validation. 
It can be seen from Table~\ref{tab:vpn} that there is consistent improvement over VSM. Thus, the learned adaptive $\beta$ acquires the ability to determine how much of the contents of memory needs to be updated based on the current new memory. $\beta$-VSM enables the memory content of different word senses to be more representative by better absorbing information from data with adaptive update, resulting in improved performance. 

\paragraph{Variation of performance with the number of senses} In order to further probe into the strengths of $\beta$-VSM, we analyze the macro F1 scores of the different models averaged over all the words in the meta-test set with a particular number of senses. In Figure \ref{fig:f1_hist}, we show a bar plot of the scores obtained from BERT for $|S| = 16$. For words with a low number of senses, the task corresponds to a higher number of effective shots and vice versa. It can be seen that the different models perform roughly the same for words with fewer senses, i.e., $2$ -- $4$. VPN is comparable to ProtoNet in its distribution of scores. But with semantic memory, VSM improves the performance on words with a higher number of senses. $\beta$-VSM further boosts the scores for such words on average. The same trend is observed for $|S| = 8$ (see Appendix \ref{sec:sense_var_8}). Therefore, the improvements of $\beta$-VSM over ProtoNet come from tasks with fewer shots, indicating that VSM is particularly effective at disambiguation in low-shot scenarios.  

\paragraph{Visualization of prototypes}
To study the distinction between the prototype distributions of word senses obtained by $\beta$-VSM, VSM and VPN, we visualize them using t-SNE~\citep{van2008visualizing}. Figure~\ref{fig:tsne_1} shows prototype distributions based on BERT for the word \textit{draw}. 
Different colored ellipses indicate the distribution of its different senses obtained from the support set. Different colored points indicate the representations of the query examples. $\beta$-VSM makes the prototypes of different word senses of the same word more distinctive and distant from each other, with less overlap, compared to the other models. Notably, the representations of query examples are closer to their corresponding prototype distribution for $\beta$-VSM, thereby resulting in improved performance. We also visualize the prototype distributions of similar vs. dissimilar senses of multiple words in Figure~\ref{fig:memory} (see Appendix \ref{sec:example_sent} for example sentences). The blue ellipse corresponds to the `set up' sense of \textit{launch} from the meta-test samples. Green and gray ellipses correspond to a similar sense of the words \textit{start} and \textit{establish} from the memory. We can see that they are close to each other. Orange and purple ellipses correspond to other senses of the words \textit{start} and \textit{establish} from the memory, and they are well separated.
For a given query word, our model is thus able to retrieve related senses from the memory and exploit them to make its word sense distribution more representative and distinctive. 


\section{Conclusion}
In this paper, we presented a model of variational semantic memory for few-shot WSD. We use a variational prototype network to model the prototype of each word sense as a distribution. 
To leverage the shared common knowledge between tasks, we incorporate semantic memory into the probabilistic model of prototypes in a hierarchical Bayesian framework. VSM is able to acquire 
long-term, general knowledge that enables learning new senses from very few examples. 
Furthermore, we propose adaptive $\beta$-VSM which learns an adaptive memory update rule from data using a lightweight hypernetwork. 
The consistent new state-of-the-art performance with three different embedding functions shows the benefit of our model in boosting few-shot WSD.

Since meaning disambiguation is central to many natural language understanding tasks, models based on semantic memory are a promising direction in NLP, more generally. Future work might investigate the role of memory in modeling meaning variation across domains and languages, as well as in tasks that integrate knowledge at different levels of linguistic hierarchy.

\bibliographystyle{acl_natbib}
\bibliography{acl2021}

\appendix

\section{Appendix}

\subsection{Implementation details} \label{sec:implementation}
In the meta-training phase, we implement $\beta$-VSM by end-to-end learning with stochastic neural networks. The inference network and hypernetwork are parameterized by a feed-forward multi-layer perceptrons (MLP). 
At meta-train time, we first extract the features of the support set via $f_{\theta}(\mathbf{x}_{\mathcal{S}})$, where $f_{\theta}$ is the feature extraction network and we use permutation-invariant instance-pooling operations to get the mean feature $\bar{f}_c^s$ of samples in the $c$-th class. Then we get the memory $M_a$ by using the support representation $\bar{f}_c^s$ of each class. 
The memory obtained $M_a$ will be fed into a small three-layers MLP network $g_{{\psi}}(\cdot)$ to calculate the mean $\boldsymbol{\mu}_{\mathbf{m}}$ and variance  $\boldsymbol{\sigma}_{\mathbf{m}}$ of the memory distribution $\mathbf{m}$, which is then used to sample the memory $\mathbf{m}$ by $\mathbf{m} \sim \mathcal{N}( \boldsymbol{\mu}_{\mathbf{m}}, diag((\boldsymbol{\sigma}_{\mathbf{m}})^2))$.
The new  memory $\bar{M}_c$ is obtained by using graph attention. The nodes of the graph are a set of feature representations of the current task samples:  $F_c = \{f_c^0, f_c^1, f_c^2, \dots, f_c^{\mathcal{N}_c}\},$ where $f_c^{\mathcal{N}_c} \in \mathbb{R}^d$, $\mathcal{N}_c = |S_c \cup Q_c|$, $f_c^0=M_c$, $f_c^{i>0} = f_{\theta}(\mathbf{x}^i_c)$. $\mathcal{N}_c$ contains all samples including both the support and query set from the $c$-th category in the current task. When we update memory, we take the new obtained memory $\bar{M}_c$ into the hypernetwork $f_{\beta}(\cdot)$ as input and output the adaptive $\beta$ to update the memory using Equation \ref{eqn:hyp_net}. We calculate the prototype of the latent distribution, i.e., the mean $\boldsymbol{\mu}_{\mathbf{z}}$ and variance $\boldsymbol{\sigma}_{\mathbf{z}}$ by another small three-layer MLP network $g_{{\phi}}(\cdot, \cdot)$, whose inputs are $\bar{f}_c^s$ and $\mathbf{m}$. Then the prototype $\mathbf{z}^{(l_{\mathbf{z}})}$ is sampled from the distribution 
$\mathbf{z}^{(l_{\mathbf{z}})} \sim \mathcal{N}( \boldsymbol{\mu}_{\mathbf{z}}, diag((\boldsymbol{\sigma}_{\mathbf{z}})^2))$.  By using the prototypical word sense of support samples and the feature embedding of query sample $\mathbf{x}_i$, we obtain the predictive value $\hat{\mathbf{y}}_i$.

At meta-test time, we feed the support representation $\bar{f}_c^s$ into the $g_{\psi}(\cdot)$ to generate the memory $\mathbf{m}_a$. Then, using the sampled memory $\mathbf{m}_a$ and the support representation $\bar{f}_c^s$, we obtain the distribution of prototypical word sense $\mathbf{z}$. Finally, we make predictions for the query sample by using the query representation extracted from embedding function and the support prototype $\mathbf{z}$.

\subsection{Hyperparameters and runtimes} \label{sec:hyperparameters}

\begin{table*}[ht]
\small
\centering
\begin{tabular}{@{}cccccccc@{}}
\toprule
\makecell{\textbf{Embedding/} \\\textbf{Encoder}} & \makecell{$|S|$} & \makecell{\textbf{Learning} \\\textbf{rate}} & \makecell{$\lambda_z$} & \makecell{$\lambda_m$} & \makecell{$L_\mathcal{z}$} & \makecell{$L_\mathbf{M}$}  \\ \midrule
\multirow{4}{*}{GloVe+GRU} 
&$4$ & $\expnumber{1}{-5}$ & $\expnumber{1}{-3}$ & $\expnumber{1}{-4}$ & 200 & 150  \\
&$8$ & $\expnumber{1}{-5}$ & $\expnumber{1}{-3}$ & $\expnumber{1}{-4}$ & 200 & 150  \\
&$16$ & $\expnumber{1}{-4}$ & $\expnumber{1}{-4}$ & $\expnumber{1}{-3}$ & 150 & 150  \\
&$32$ & $\expnumber{1}{-4}$ & $\expnumber{1}{-3}$ & $\expnumber{1}{-3}$ & 150 & 150  \\ \midrule
\multirow{4}{*}{ELMo+MLP} 
&$4$ & $\expnumber{1}{-5}$ & $\expnumber{1}{-4}$ & $\expnumber{1}{-4}$ & 200 & 150  \\
&$8$ & $\expnumber{1}{-5}$ & $\expnumber{1}{-4}$ & $\expnumber{1}{-4}$ & 200 & 150  \\
&$16$ & $\expnumber{1}{-4}$ & $\expnumber{1}{-3}$ & $\expnumber{1}{-3}$ & 150 & 150  \\
&$32$ & $\expnumber{1}{-4}$ & $\expnumber{1}{-3}$ & $\expnumber{1}{-3}$ & 150 & 150  \\ \midrule
\multirow{4}{*}{BERT} 
&$4$ & $\expnumber{5}{-6}$ & $\expnumber{1}{-3}$ & $\expnumber{1}{-4}$  & 200 & 200  \\
&$8$ & $\expnumber{5}{-6}$ & $\expnumber{1}{-3}$ & $\expnumber{1}{-4}$ & 200 & 200 \\
&$16$ & $\expnumber{1}{-6}$ & $\expnumber{1}{-4}$ & $\expnumber{1}{-4}$ & 150 & 150  \\
&$32$ & $\expnumber{1}{-4}$ & $\expnumber{1}{-3}$ & $\expnumber{1}{-4}$ & 150 & 100  \\  
\bottomrule
\end{tabular}
\caption{Hyperparameters used for training the models.}
\label{tab:hyperparams}
\end{table*}

We present our hyperparameters in Table~\ref{tab:hyperparams}.
For Monte Carlo sampling, we set different $L_Z$ and $L_M$ for the each embedding function and $|S|$, which are chosen using the validation set.
Training time differs for different $|S|$ and different embedding functions. Here we give the training time per epoch for $|S| = 16$. For GloVe+GRU, the approximate training time per epoch is $20$ minutes; for ELMo+MLP it is $80$ minutes; and for BERT, it is $60$ minutes. The number of meta-learned parameters for GloVe+GRU is $\theta$ are $889, 920$; for ELMo+MLP it is $262, 404$; and for BERT it is $\theta$ are $107, 867, 328$. 
We implemented all models using the PyTorch framework and trained them on an NVIDIA Tesla V100.

\subsection{Variation of performance with the number of senses} \label{sec:sense_var_8}

\begin{figure*}[t]
    \centering
    \begin{subfigure}[b]{0.24\textwidth}
         \centering
         \includegraphics[width=\textwidth]{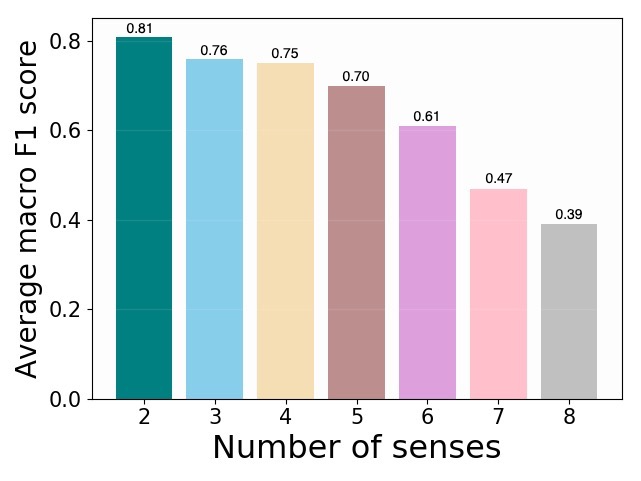}
         \caption{ProtoNet}
     \end{subfigure}
     \hfill
    \begin{subfigure}[b]{0.24\textwidth}
         \centering
         \includegraphics[width=\textwidth]{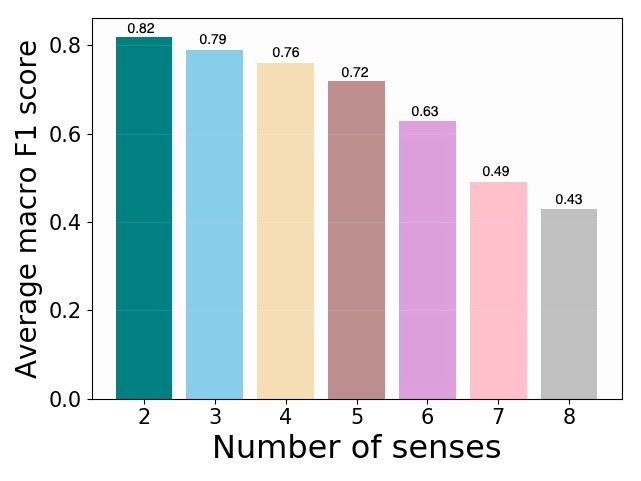}
         \caption{VPN}
     \end{subfigure}
     \hfill
     \begin{subfigure}[b]{0.24\textwidth}
         \centering
         \includegraphics[width=\textwidth]{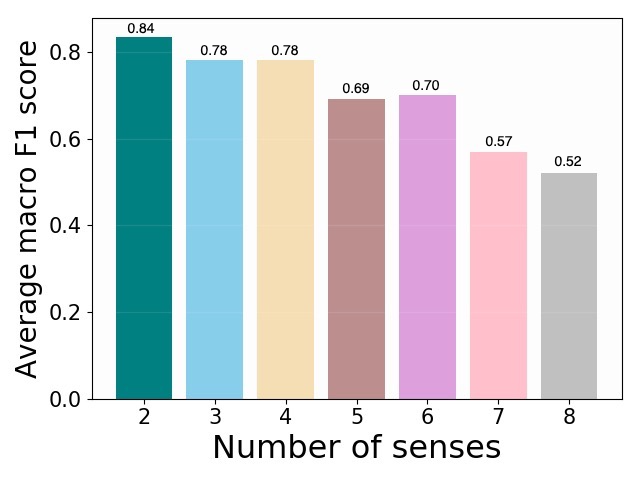}
         \caption{VSM}
     \end{subfigure}
     \hfill
     \begin{subfigure}[b]{0.24\textwidth}
         \centering
         \includegraphics[width=\textwidth]{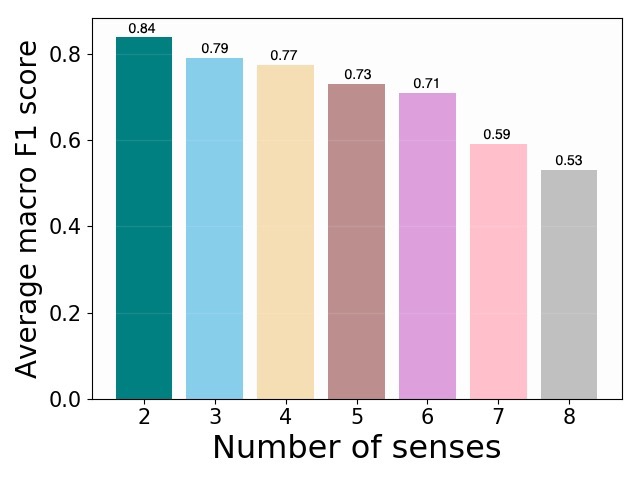}
         \caption{$\beta$-VSM}
     \end{subfigure}
    \caption{Distribution of average macro F1 scores over number of senses for BERT-based models with $|S| = 8$.}
    \label{fig:f1_hist_8}
\end{figure*}

To further demonstrate that $\beta$-VSM achieves better performance in extremely data scarce scenarios, we also analyze variation of macro F1 scores with the number of senses for BERT and $|S| = 8$. In Figure~\ref{fig:f1_hist_8}, we observe a similar trend as with $|S| = 16$. $\beta$-VSM has an improved performance for words with many senses, which corresponds to a low-shot scenario. For example, with $8$ senses, the task is essentially one-shot.

\subsection{Example sentences to visualize prototypes} \label{sec:example_sent}
In Table \ref{tab:example_sentences}, we provide some example sentences used to generate the plots in Figure 4. These examples correspond to words \textit{launch}, \textit{start} and \textit{establish}, and contain senses `set up', `begin' and `build up'.

\begin{table*}[ht!]
\centering
\begin{tabular}{p{1.5cm} p{1.5cm} p{11cm}}
\toprule
\textbf{Word} & \textbf{Sense} & \textbf{Sentence} \\ \midrule
launch        & set up         & The Corinthian Yacht Club in Tiburon \underline{launches} its winter races Nov. 5.  \\
launch        & set up         & The most infamous of all was \underline{launched} by the explosion of the island of Krakatoa in 1883; it raced across the Pacific at 300 miles an hour devastated the coasts of Java and Sumatra with waves 100 to 130 feet high, and pounded the shore as far away as San Francisco. \\
launch        & set up         & In several significant cases, such as India, a decade of concentrated effort can \underline{launch} these countries into a stage in which they can carry forward their own economic and social progress with little or no government-to-government assistance.                        \\
start         & set up         & With these maps completed, the inventory phase of the plan has been \underline{started}. \\
start         & begin          & Congress \underline{starts} another week tomorrow with sharply contrasting forecasts for the two chambers. \\
establish     & set up         & For the convenience of guests bundle centers have been \underline{established} throughout the city and suburbs where the donations may be deposited between now and the date of the big event. \\
establish     & build up       & From the outset of his first term, he \underline{established} himself as one of the guiding spirits of the House of Delegates. \\
\bottomrule
\end{tabular}
\caption{Example sentences for different word-sense pairs used to generate the visualization in Figure \ref{fig:memory}.}
\label{tab:example_sentences}
\end{table*}

\subsection{Results on the meta-validation set}
We provide the results on the on the meta-validation set in the Table~\ref{tab:val}, to better facilitate reproducibility. 

\begin{table*}[ht]
\small
\centering
\begin{tabular}{llllll}
\toprule
\multirow{2}{*}{\makecell{\textbf{Embedding/} \\ \textbf{Encoder}}} & \multicolumn{1}{c}{\multirow{2}{*}{\textbf{Method}}} &  \multicolumn{4}{c}{\textbf{Average macro F1 score}} \\
                       & & \multicolumn{1}{c}{$|S| = 4$} & \multicolumn{1}{c}{$|S| = 8$}  & \multicolumn{1}{c}{$|S| = 16$} & \multicolumn{1}{c}{$|S| = 32$} \\ \midrule
\multirow{4}{*}{GloVe+GRU} 
& ProtoNet & 0.591 $\pm$ 0.008 & 0.615 $\pm$ 0.001 & 0.638 $\pm$ 0.007 & 0.634 $\pm$ 0.006 \\
& VPN& 0.602 $\pm$ 0.004 & 0.624 $\pm$ 0.004 & 0.646 $\pm$ 0.006 & 0.651 $\pm$ 0.005 \\
& VSM& 0.617 $\pm$ 0.005 & 0.635 $\pm$ 0.005 & 0.649 $\pm$ 0.004 & 0.673 $\pm$ 0.006 \\
& $\beta$-VSM  & \textbf{0.622 $\pm$ 0.005} & \textbf{0.649 $\pm$ 0.004} & \textbf{0.657 $\pm$ 0.005} & \textbf{0.680 $\pm$ 0.006} \\
\midrule
\multirow{4}{*}{ELMo+MLP} 
& ProtoNet & 0.682 $\pm$ 0.008 & 0.701 $\pm$ 0.007 & 0.741 $\pm$ 0.007 & 0.722 $\pm$ 0.011 \\
& VPN & 0.689 $\pm$ 0.004 & 0.709 $\pm$ 0.006 & 0.749 $\pm$ 0.005 & 0.748 $\pm$ 0.004  \\ 
& VSM& 0.693 $\pm$ 0.005 & 0.712 $\pm$ 0.007 & 0.754 $\pm$ 0.006 & 0.755 $\pm$ 0.006  \\ 
& $\beta$-VSM & \textbf{0.701 $\pm$ 0.006} & \textbf{0.723 $\pm$ 0.005} & \textbf{0.760 $\pm$ 0.005} & \textbf{0.761 $\pm$ 0.004}  \\
\midrule
\multirow{4}{*}{BERT} 
& ProtoNet & 0.742 $\pm$ 0.007 & 0.759 $\pm$ 0.013 & 0.786 $\pm$ 0.004 & 0.770 $\pm$ 0.009 \\
& VPN & 0.752 $\pm$ 0.011 & 0.769 $\pm$ 0.005 & 0.793 $\pm$ 0.003 & 0.785 $\pm$ 0.004 \\
& VSM& 0.767 $\pm$ 0.009 & 0.778 $\pm$ 0.005 & 0.801 $\pm$ 0.006 & 0.815 $\pm$ 0.005 \\
& $\beta$-VSM  & \textbf{0.771 $\pm$ 0.008} & \textbf{0.784 $\pm$ 0.006} & \textbf{0.810 $\pm$ 0.004} & \textbf{0.829 $\pm$ 0.004} \\
\bottomrule                       
\end{tabular}
\caption{Average macro F1 scores of the meta-validation words.}
\label{tab:val}
\end{table*}

\end{document}